\PassOptionsToPackage{colorlinks=true,allcolors=gray}{hyperref}

 \documentclass[accepted]{uai2023} 

\usepackage[american]{babel}

\usepackage{multicol}

\usepackage{natbib} 
\usepackage{mathtools} 
\usepackage{booktabs} 
\usepackage{tikz} 
\usepackage{xspace}
\usepackage{siunitx}

\usepackage[noend]{algpseudocode} 
\usepackage[ruled,vlined,linesnumbered]{algorithm2e}

\SetCommentSty{mycommfont}
\SetKwProg{Fct}{Fct}{}{}
\usepackage{amsfonts}
\usepackage{amsthm,mathtools}
\allowdisplaybreaks

\newtheorem{definition}{Definition}

\newcommand{\eqdef}     {\stackrel{{\textrm{\rm\tiny def}}}{=}}

\def\ie{{\em i.e.}\xspace}
\def\eg{{\em e.g.}\xspace}
\def\cf{{\em cf.}\xspace}

\def\reals{{\mathbb R}}
\def\cS{{\cal S}}
\def\cA{{\cal A}}

\def\cI{{\cal I}}
\def\cZ{{\cal Z}}
\def\cE{{\cal E}}
\def\va{{\mathbf a}}
\def\vo{{\mathbf o}}
\def\vn{{\mathbf n}}
\def\nNI{\#ni} 
\def\fsc{\mathit{fsc}} 
\newcommand{\infJESP}[1][]{Inf-JESP{#1}\xspace}
\usepackage[noabbrev]{cleveref}
\DeclareRobustCommand{\abbrevcrefs}{%
\Crefname{appendix}{App.}{Apps.}%
\Crefname{section}{Sec.}{Secs.}%
\Crefname{equation}{Eq.}{Eqs.}%
\Crefname{figure}{Fig.}{Figs.}%
\Crefname{algorithm}{Alg.}{Algs.}%
\Crefname{tabular}{Tab.}{Tabs.}%
\Crefname{lemma}{Lem.}{Lems.}%
\Crefname{corollary}{Cor.}{Cors.}%
\Crefname{theorem}{Thm.}{Thms.}%
\Crefname{proposition}{Prop.}{Props.}%
\Crefname{line}{L.}{Ls.}%
\crefname{appendix}{app.}{apps.}%
\crefname{section}{sec.}{secs.}%
\crefname{equation}{eq.}{eqs.}%
\crefname{figure}{fig.}{figs.}%
\crefname{algorithm}{alg.}{algs.}%
\crefname{tabular}{tab.}{tabs.}%
\crefname{lemma}{lem.}{lems.}%
\crefname{corollary}{cor.}{cors.}%
\crefname{theorem}{thm.}{thms.}%
\crefname{proposition}{prop.}{props.}%
\crefname{line}{l.}{ls.}%
}

\DeclareRobustCommand{\Cshref}[1]{{\abbrevcrefs\Cref{#1}}}
\mathchardef\mhyphen="2D
\crefalias{AlgoLine}{line}%
\newcommand\linesrefAnd[2]{lines~\ref{#1} and \ref{#2}}
\newcommand\linesref[2]{lines~\ref{#1}--\ref{#2}}

\usepackage[normalem]{ulem}

\usepackage{xcolor}

\title{  Monte-Carlo Search for an Equilibrium in Dec-POMDPs}

\author[1]{Yang You}
\author[1]{Vincent Thomas}
\author[1]{Francis Colas}
\author[1]{Olivier Buffet}
\affil[1]{%
  Université de Lorraine, INRIA, CNRS, LORIA,  
  Nancy, France
}

\begin{document}
\maketitle


\begin{abstract}
  Decentralized partially observable Markov decision processes (Dec-POMDPs) formalize the problem of designing individual controllers for a group of collaborative agents under stochastic dynamics and partial observability.
  Seeking a global optimum is difficult (NEXP complete), but seeking a Nash equilibrium ---each agent policy being a best response to the other agents--- is more accessible, and allowed addressing infinite-horizon problems with solutions in the form of finite state controllers.
  In this paper, we show that this approach can be adapted to cases where only a generative model (a simulator) of the Dec-POMDP is available.
  This requires relying on a simulation-based POMDP solver to construct an agent's FSC node by node.
  A related process is used to heuristically derive initial FSCs. 
  Experiment with benchmarks shows that MC-JESP is competitive with exisiting Dec-POMDP solvers, even better than many offline methods using explicit models.
\end{abstract}

\section{Introduction}\label{sec:intro}


%
The framework of Decentralized Partially Observable Markov Decision Processes (Dec-POMDPs) allow modeling collaborative multi-agent systems, the objective being to equip them with individual policies that maximize some common performance criterion.
%
%
However, solving Dec-POMDPs is challenging since the environment evolves according to all agent's actions,
and each agent performs its action only based on its local action-observation histories.
To ensure finding a global optimum, one thus needs to reason about all individual policies together.
As a consequence of this interdependency, even for a finite-horizon Dec-POMDP, the solving process has been proven to be NEXP in the worst case \citep{Bernstein02}, and solving an infinite-horizon Dec-POMDP is undecidable \citep{MADANI20035, 10.5555/2967142}.

\citeauthor{JESP} propose an alternative approach called JESP (Joint Equilibrium-Based Search for Policies) \citep{JESP},  which avoids this interdependency in the solving process by searching for a Nash equilibrium, \ie, each agent's policy is a best response to other agents' policies.
JESP operates an iterative optimization process over each agent.
In each iteration, it builds agent $i$'s best-response policy considering other agents' policies are fixed.
%
%
A Nash equilibrium is therefore guaranteed when no further improvement is possible.
In JESP, each agent's policy is represented in a tree structure, which limits its usage only to finite-horizon problems.
To overcome this limitation, \infJESP (Infinite-Horizon JESP) \citep{InfJESP} extends JESP to infinite-horizon Dec-POMDPs by representing each agent policy as a finite-state controller (FSC).
Two advantages of \infJESP are that
\begin{enumerate*}
\item it often achieves near-global optima despite only searching for local ones, and
\item its FSCs make for interpretable policies if their size is reasonable.
\end{enumerate*}
However, both methods require an explicit Dec-POMDP model which details the exact environment dynamics.

In this paper, we propose a new algorithm called MC-JESP (Monte-Carlo Joint Equilibrium-based Search for Policies),
%
which is a simulation-based version of \infJESP.
%
%
In each iteration, MC-JESP builds an agent's FSC node by node using a Monte-Carlo (POMDP) planner relying on a black-box Dec-POMDP simulator, along with the other agents' FSCs.
Experiments show that MC-JESP is competitive with state-of-the-art infinite-horizon Dec-POMDP solvers based either on exact or generative models.
%

The structure of this paper is organized as follows:
\Cref{sec:related_work} discusses related work on solving Dec-POMDPs.
\Cshref{sec:background} gives background about Dec-POMDPs, POMDPs, FSCs, and \infJESP.
Our contributions are presented in \Cshref{sec:MCJESP}, and experiments with comparisons to state-of-the-art Dec-POMDP solvers in \Cshref{sec:exp}.
Finally, we conclude this work in \Cshref{sec:conclusion} and discuss future perspectives.

\section{Related Work}
\label{sec:related_work}

Recently, there has been significant progress in infinite-horizon Dec-POMDP planning, and state-of-the-art methods fall into three main types.
The first type of methods estimates the best parameters of finite-state controllers (FSCs) of each agent \citep{AmaBerZil-jaamas10}, and addresses Dec-POMDPs as an inference problem via
Expectation-Maximization methods \citep{PajPel-ijcai11,PajPel-nips11, kumar2012, KumZilTou-jair15}.
The second type consists in transforming the Dec-POMDP problem into a Markov decision process with a state space of sufficient statistics \citep{MacIsb-nips13, dibangoye2014error, DibAmaBufCha-jair16}.
The third type searches for Nash equilibrium solutions, \ie, each agent's policy being a best response to the other agents' policies \citep{JESP,BerHanZil-ijcai05, InfJESP}.
%

However, for large problems or real applications,  it may be challenging to represent the system's dynamics explicitly.
Often, only a black-box simulator (also called a generative model) is available.
%
%
Although the algorithms mentioned previously with explicit models cannot be directly applied, most state-of-the-art simulation-based methods are inspired by them.
For example,
\citet{Wu2013} propose to use a Monte-Carlo Expectation Maximization (MCEM) for estimating the parameters of agents' FSCs with generative models.
\citet{Liu2015} improve MCEM by constructing agent FSCs using the stick-breaking prior and allowing a variable FSC size.
%
%
On the other hand, similar to FB-HSVI \citep{DibAmaBufCha-jair16} (which uses explicit models), the simulation-based method oSARSA \citep{pmlr-v80-dibangoye18a} focuses on recasting Dec-POMDPs into occupancy-state MDPs, where each occupancy-state is a sufficient statistics.

Last but not least, some multi-agent reinforcement learning (MARL) algorithms are also interested in solving Dec-POMDPs with black-box simulators.
However, most of them \citep{VDN2017, rashid2018qmix, son2019qtran, rashid2020weighted} 
have not been evaluated on classical Dec-POMDP benchmarks \citep{Seuken2007ImprovedMD, amato2009achieving}.
Only a few MARL algorithms conducted experiments on such domains but were limited to finite-horizon settings \citep{lee2019improved}, or failed to obtain state-of-the-art results \citep{KRAEMER201682}.

\section{Background}
\label{sec:background}

\subsection{Dec-POMDP}


The problem of finding optimal collaborative behaviors for a group of
agents under stochastic dynamics and partial observability is
typically formalized as a {\em decentralized partially observable
  Markov decision process} (Dec-POMDP).

\begin{definition}
  A {\em Dec-POMDP} with $|\cI|$ agents is represented as a tuple $M \equiv \langle \cI, \cS, \cA, \Omega, T, O, R, b_0, H, \gamma \rangle$, where:
  $\cI = \{1, \dots, |\cI|\}$ is a finite set of {\em agents};
  $\cS$ is a finite set of {\em states};
  $\cA = \bigtimes_i \cA^i$ is the finite set of joint actions, with %
  $\cA^i$ the set of agent $i$'s {\em actions}; %
  $\Omega = \bigtimes_i \Omega^i$ is the finite set of joint
  observations, with %
  $\Omega^i$ the set of agent $i$'s {\em observations}; %
  $T: \cS \times \cA \times \cS \to \reals$ is the {\em transition
    function}, with %
  $T(s,\va,s')$ the probability of transiting from $s$ to $s'$ if $\va$ is
  performed;
  $O: \cA \times \cS \times \Omega \to \reals$ is the
  {\em observation function}, with %
  $O(\va,s',\vo)$ the probability of observing $\vo$ if $\va$ is performed and
  the next state is $s'$;
  $R: \cS \times \cA \to \mathbb{R}$ is the {\em reward function},
  with %
  $R(s,\va)$ the immediate reward for executing $\va$ in $s$;
  $b_0$ is the {\em initial probability distribution} over states;
  $H \in \mathbb{N} \cup \{\infty\}$ is the (possibly infinite) {\em time horizon};
  $\gamma \in [0,1)$ is the {\em discount factor} applied to future rewards.
\end{definition}

An agent's $i$ action {\em policy} $\pi^i$ maps its possible
action-observation histories to actions.
The objective is then to find a joint policy
$\pi=\langle \pi^1, \dots, \pi^{|\cI|} \rangle$ that maximizes the expected discounted return from $b_0$:
\begin{align*}
  V^{\pi}_H(b_0)
  & \eqdef \mathbb{E}\left[ \sum_{t=0}^{H-1} \gamma^{-t} r(S_t, A_t) \mid S_0 \sim b_0, \pi \right].
\end{align*}

However, we often do not know the exact transition, observation, and reward functions for large problems or real-world applications,
%
but may rely on a generative model (black-box simulator) $G$, which, given a state-action pair $\langle s,\va \rangle$, samples a triplet $\langle s',\vo,r \rangle$.

\subsection{POMDP}
%
%
%
%
In this work, we will consider one agent $i$ at a time,
%
%
and thus end up solving a single-agent partially observable Markov decision problem (POMDP) in each iteration, \ie, the particular case of a single-agent Dec-POMDP ($\cI=\{1\}$).
%
In a POMDP, an optimal policy $\pi^*$ exists whose input is the belief state
$b$, \ie, the probability distribution over states given the current action-observation history.
For finite $H$, the optimal value function (which allows deriving
$\pi^*$) is recursively defined as:
\begin{align*}
  V^*_h(b)
  & = \max_a \left[r(b, a) + \gamma \sum_{o} Pr(o \mid b, a) V^*_{h-1}(b^{a,o}) \right],
\end{align*}
where %
(i) $r(b,a) = \sum_s b(s)\cdot r(s,a)$; %
(ii) $Pr(o \mid b, a)$ depends on the dynamics; and %
(iii) $b^{a,o}$ is the belief updated upon performing $a$ and
perceiving $o$.
%
%

\subsection{Finite State Controllers}
\label{sec:FSC_def}

In POMDPs as in Dec-POMDPs, solution policies can also be sought for in
the form of {\em finite state controllers} (FSC) (also called {\em
  policy graphs} \citep{MeuKimKaeCas-uai99}), \ie, automata whose
transitions from one internal state to the next depend on the received
observations and generate the actions to be performed.

\begin{definition}
  For some POMDP sets $\cA$ and $\Omega$, %
  a (deterministic) {\em FSC} is specified by a tuple
  $\fsc \equiv \langle N, \eta, \psi \rangle$, where:
  \begin{itemize}
  \item $N$ is a finite set of nodes, %
    with $n_0$ the start node; %
  \item $\eta: N \times \Omega \to N $ is the node transition function; %
    $n'=\eta(n,o)$ is the next node and observing $o$ from node $n$;
  \item $\psi: N \to \cA $ is the action-selection function of the FSC; %
    $a=\psi(n)$ is the action triggered when in node $n$.
  \end{itemize}
\end{definition}



\subsection{Solving Dec-POMDPs by finding Nash Equilibria (Infinite-Horizon JESP)}
\label{sec:infJESP}

%
%
%
%
\infJESP (Infinite-Horizon JESP) \citep{InfJESP} is an infinite-horizon Dec-POMDP solver, which is based on \citeauthor{JESP}'s JESP [\citeyear{JESP}], but replaces the policy tree representation by a finite-state controller (FSC) for each agent's policy.
This modification allows solving infinite-horizon problems rather than finite-horizon ones, and %
may help scaling up to larger problems.
%
%
More specifically, in \infJESP (\Cref{alg:JESP_main}), each iteration derives (\cref{codes:BuildModel_infJESP}) the explicit model of a (best-response) POMDP for agent $i$ by fixing the other agents' FSCs (index ``$\neq i$'') and using an extended state space $e^t \in \cE$, \ie, containing:
\begin{itemize}
 \item $s^t$, the current state of the Dec-POMDP problem,
 \item $\vn_{\neq i}^{t} \equiv \langle n_j^{t} \rangle_{j\neq i}$, the current nodes of other agents, and
 \item $\tilde o_{i}^{t} $, agent $i$'s current observation.
\end{itemize}
Denoting $\eta_{\neq i}(\vn_{\neq i}^{t}, \vo^{t+1}_{\neq i}) = \langle \eta(n_j^{t}, \tilde o_j^{t+1}) \rangle_{j\neq i} $ and $\psi_{\neq i}(\vn_{\neq i}^{t}) = \langle \psi_j(n_j^{t}) \rangle_{j\neq i} $,
this leads to a valid POMDP with the following dynamics:\footnote{Note: \citet{InfJESP} provide formulas for stochastic FSCs.}
\begin{align*}
  &  T_e(e^t, a^t_i, e^{t+1})  = Pr(e^{t+1}| e^t, a^t_i) \\
  & \quad = \sum_{\vo^{t+1}_{\neq i}}T(s^t, \langle  \psi_{\neq i}(\vn_{\neq i}^{t}), a^t_{i} \rangle, s^{t+1})
  \cdot \mathbf{1}_{ \vn_{\neq i}^{t+1} = \eta_{\neq i}(\vn_{\neq i}^{t}, \vo^{t+1}_{\neq i}) } \\
  & \qquad \cdot O(s^{t+1}, \langle \psi_{\neq i}(\vn_{\neq i}^{t}), a^t_{i} \rangle, \langle \vo^{t+1}_{\neq i}, o^{t+1}_{i} \rangle),
  \\ %
  & O_e(a^t_i, e^{t+1}_i, o^{t+1}_i)
  = Pr( o^{t+1}_i | a^t_i, e^{t+1}_i) \\
  & \quad = Pr( o^{t+1}_i | a^t_i, \langle s^{t+1}, \vn^{t+1}_{\neq i}, \tilde o^{t+1}_i \rangle) %
  = \mathbf{1}_{o^{t+1}_i = \tilde  o^{t+1}_i},
  \\
  & r_e(e^t, a^t_i) =  r(s^t, a^t_i, \psi_{\neq i}(\vn_{\neq i}^{t}) ).
\end{align*}
Then, \infJESP solves this explicit POMDP for agent $i$ using an $\epsilon$-optimal offline POMDP solver (SARSOP \citep{sarsop}) and derives an FSC $\fsc'_i$ that approximates the solution policy (cf. \cref{codes:getFSC}, which does not distinguish both steps).
$\fsc_i'$ is then evaluated (\cref{codes:PolicyEval_infJESP}) and retained only if it improves on $i$'s previous FSC, $\fsc_i$, so that \infJESP stops when an approximate Nash equilibrium is obtained, which is detected using a counter $\nNI$.

\begin{algorithm}
  \caption{Inf-JESP's Local Search}
  \label{alg:JESP_main}
  \DontPrintSemicolon
  \SetInd{.3em}{.6em}

  \SetKwFunction{LocalSearch}{{\bf LocalSearch}}
  \SetKwFunction{ComputeFSC}{{\bf ComputeFSC}}
  \SetKwFunction{getBRpomdpModel}{{\bf getBRpomdp}}
  \SetKwFunction{Eval}{{\bf Eval}}

  [Input:] %
  $b^0$: initial belief $\mid$ %
  $M$: Dec-POMDP model $\mid$ %
  \linebreak %
  $\fsc$: initial FSCs \; %
  \Fct{\LocalSearch{$b_0, M, \fsc$} }{
    $v_{bestL} \gets eval(\fsc)$ \;
    $\nNI \gets 0$   \tcp*[h]{\#(iterations w/o improvement)} \;
    $i\gets 1$  \tcp*[h]{Id of current agent} \;
    \Repeat( \tcp*[h]{Cycle over agents} ){
      $\nNI=|\cI|$ \tcp*[h]{No improvement in last cycle.}
    }{
      $b^0_{BR}, M_{BR} \gets \getBRpomdpModel(b^0, M, \fsc_{\neq i})$ \label{codes:BuildModel_infJESP} \;
      $\fsc'_i \gets $\ComputeFSC{$b^0_{BR}, M_{BR}$ } \label{codes:getFSC} \;
      $v \gets \Eval(\fsc'_i,  b^0_{BR},  M_{BR})$ \label{codes:PolicyEval_infJESP} \;
      \uIf(\tcp*[h]{Keep new FSC if better}){$v > v_{bestL}$}{
        $\fsc_i \gets \fsc'_i$\;
        $v_{bestL} \gets v$\;
        $\nNI \gets 0$\;
      }
      \Else(\tcp*[h]{increment $\nNI$}){
        $\nNI \gets \nNI+1$\;
      }
      $i \gets (i \mod |\cI|) + 1$\;
    }
    \Return{$\langle \fsc, v_{bestL} \rangle$}
  }
\end{algorithm}

In practice (see \citep{InfJESP} or \Cshref{sec:results}), this search for an equilibrium often allows finding near-global optima either using some random restarts, or initial FSCs obtained through relaxing the Dec-POMDP.

\section{Monte-Carlo JESP}
\label{sec:MCJESP}

%
%
%

As \infJESP, we aim to find Nash equilibrium infinite-horizon solutions by iteratively building each agent's best response to other agents' fixed policies.
We thus stick to representing policies as FSCs, and to using the same algorithmic scheme for the local search as presented in \Cshref{alg:JESP_main}.

This requires relying on the same best-response POMDPs, \ie, in particular with the same extended state $s^t_{e} = \langle s^t, \vn_{\neq i}^{t}, o_{i}^{t}  \rangle$.
However, lacking an explicit Dec-POMDP model, we cannot derive explicit POMDP models.
%
%
To address this issue, in MC-JESP, we propose an alternative approach that relies on \textit{best-response generative POMDP models}  (noted $G_{BR}$) derived from the Dec-POMDP simulator and other agents' FSCs.
%
In the following, we discuss %
how to build such models, how to derive solution FSCs, %
how to obtain initial heuristic FSCs, and %
what are the
properties of the resulting approach.

\subsection{Best-Response Generative Model}
\label{sec:BR_generative_model}

\begin{figure}
     \centering
     \includegraphics[width=1.0 \linewidth]{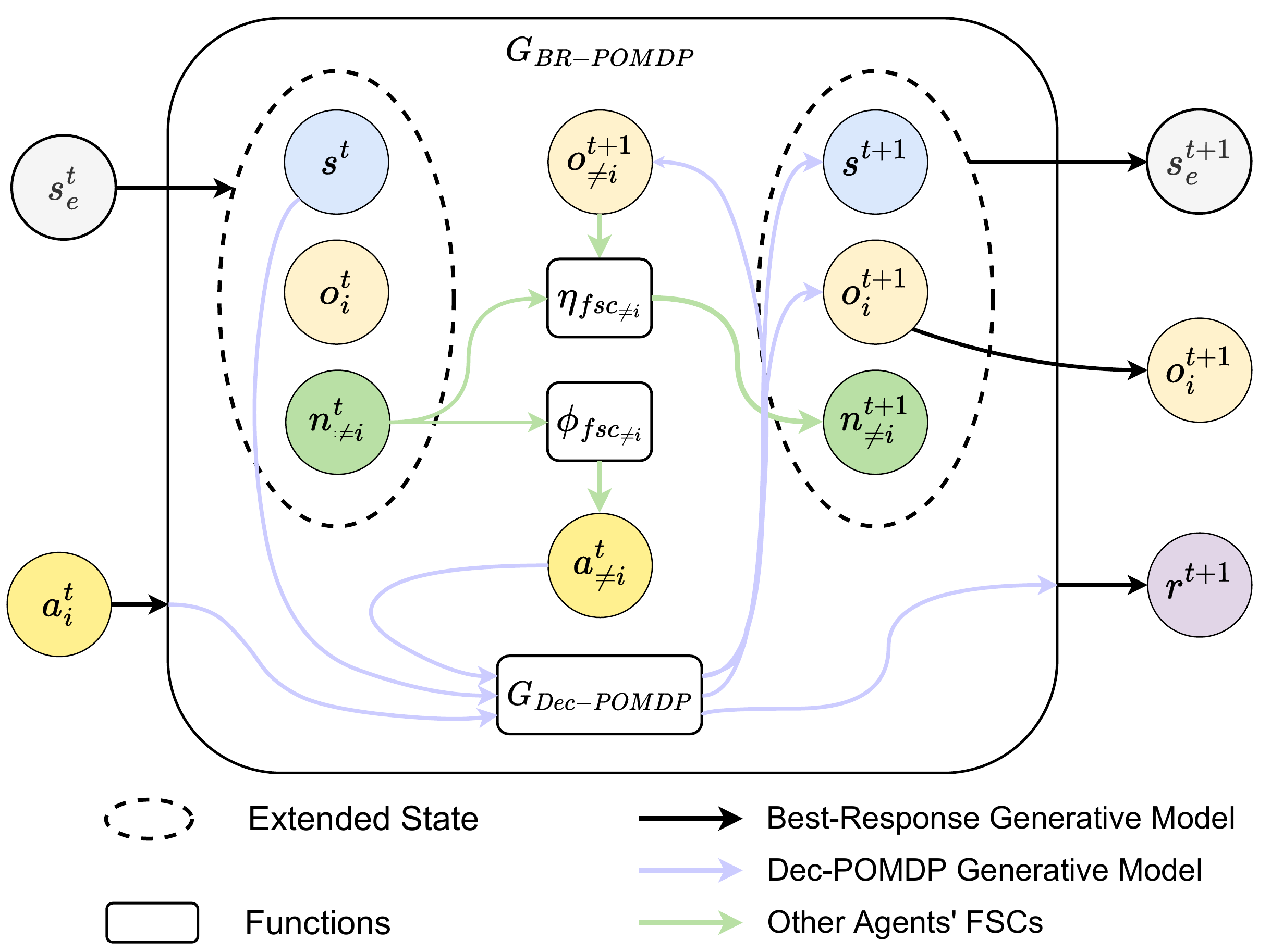}
     \caption{Structure of the best-response POMDP generative model $G_{BR}$,
       with inputs and outputs represented as:
       blue arrows for the Dec-POMDP simulator $G$;
       green arrows for agents $\neq i$' FSCs; and
       black arrows for $G_{BR}$.
     }
  \label{fig:BRGenerativeModel}
\end{figure}

A generative POMDP model $G_{BR}$ for agent $i$ has to sample the next extended state $s^{t+1}_e$, observation $o^{t+1}_i$, and reward $r^{t+1}$, given a current extended state $s^t_e$ and action $a^t_i$.
As illustrated in \Cref{fig:BRGenerativeModel} and as detailed in \Cshref{alg:MCJESP_Extended_G}, this can be achieved by relying only on the Dec-POMDP simulator and the other agents' FSCs.
The algorithm first decomposes the extended state, and gets other agents' actions $\va^t_{\neq i}$ according to action-selection functions $\psi_{\neq i} \equiv \langle \psi_j \rangle_{j\neq i}$  (\cref{line:decompose_extended_state,line:GetOthersAction}).
Then, in \cref{line:MC_JESP_step_of_G}, the joint action $ \langle a^t_{i}, \va^t_{\neq i} \rangle$ is passed to the Dec-POMDP simulator $G$, which outputs the next state $s^{t+1}$, joint observation $\langle o^{t+1}_i, \vo^{t+1}_{\neq i}  \rangle$ and instant reward $r^{t+1}$.
With the other agents' observations $\vo^{t+1}_{\neq i}$, \cref{line: GetOthersNextNode} computes their next nodes $\vn^{t+1}_{\neq i} \equiv \langle n^{t+1}_j \rangle_{j\neq i}$.
In the end, we build the next extended state $s^{t+1}_e$ and return the results (\cref{line:built_next_ste,line:MCJESP_return}).
In this algorithm, stochasticity exists only in the Dec-POMDP simulator $G$, the FSC functions $\psi$ and $\eta$ being deterministic.


\newcommand{\tmp}[1]{
  \phantom{$\langle s^t, \vn^t_{\neq i}, o^t_i \rangle$}\makebox[0pt][r]{#1}
}

\begin{algorithm}
  \caption{$i$'s Best-Response Generative Model} \label{alg:MCJESP_Extended_G}
  \DontPrintSemicolon
  \SetInd{.3em}{.6em}

  \SetKwFunction{ProcessAction}{{\bf getNxtBeliefs}}

  {[Input:]} %
  $s^t_e$: extended state $\mid$ %
  $a^t_i$: agent $i$'s action \; 
  {[Parameters:]} %
  $G$: Dec-POMDP simulator $\mid$ \linebreak %
  $\fsc_{\neq_i} \equiv \langle  N_{\neq i}, \psi_{\neq i}, \eta_{\neq i} \rangle$: other agents' FSCs \\
  \Fct{\bf $G_{BR}(s^t_e,a^t_i, [G, \fsc_{\neq_i}])$}{
    \tmp{$\langle s^t, \vn^t_{\neq i}, o^t_i \rangle$} $\gets s^t_{e} $ \label{line:decompose_extended_state}
    \tcp*[f]{extract $s^t_e$'s 3 components} \;
    \tmp{$\va^t_{\neq i}$} $\gets \psi_{\neq i}(\vn^t_{\neq i})$   \label{line:GetOthersAction}
    \tcp*[f]{get action from FSC} \;
    $s^{t+1}, \vo^{t+1}, r^{t+1} \gets G(s^t, \va^t )$ \label{line:MC_JESP_step_of_G}
    \tcp*[f]{sample transition} \;
    \tmp{$\vn^{t+1}_{\neq i}$} $\gets \eta(\vn^t_{\neq i}, \vo^{t+1}_{\neq i})$ \label{line: GetOthersNextNode}
    \tcp*[f]{evolve FSC} \;
    \tmp{$s^{t+1}_{e}$} $\gets \langle s^{t+1}, \vn^{t+1}_{\neq i}, o^{t+1}_i  \rangle$ \label{line:built_next_ste}
    \tcp*[f]{build $s^{t+1}_e$} \;
    \textbf{return} $s^{t+1}_{e}, o^{t+1}_i , r^{t+1}$ \label{line:MCJESP_return} \tcp*[f]{return step results}
  }
\end{algorithm}

\subsection{Computing Agent $i$'s FSC using Monte-Carlo methods}
\label{sec:Compute_FSC}

In the previous section, we demonstrate how to build the best-response generative model $G_{BR}$ for agent $i$ considering others' fixed FSCs.
However, unlike in \infJESP, state-of-the-art point-based POMDP solvers (see, \eg, \citep{PBVI,Smith_2004_HSVI,sarsop}) require exact models, and thus can not be used in MC-JESP.
%
%
%
We thus rely on a simulation-based solver, \ie, POMCP \citep{NIPS2010_edfbe1af}, which is an online algorithm, \ie, it focuses on returning the best action for the current belief.
Therefore, the question is how to use a simulator ($G_{BR}$) and an online simulation-based solver to obtain agent $i$'s FSC.
To answer it, we propose an algorithm that
\begin{enumerate*}
\item uses this Monte-Carlo planner (POMCP) to compute the best action for a  given FSC node, which is labeled by a unique belief; and
\item expands reachable beliefs, \ie, creates new FSC nodes using computed actions to gradually build a complete FSC.
\end{enumerate*}
%
Moreover, to control the computational cost, we explicitly bound the FSC size with a given parameter $N_{max\mhyphen\fsc} \in \mathbb{N}$.

In the proposed algorithm, each FSC node is attached to
\begin{enumerate*}
\item an approximate belief $b$ (with at least $N_{min\mhyphen part}$ particles),
\item a preferred action $a_i$, and
\item a weight $w$ that estimates the probability to reach that node at least once during execution.
\end{enumerate*}
%
%
%
As detailed in \Cref{alg:MCJESP_BuildFSC}, this information is first gathered for initial belief $b_0$ by calling
POMCP to get agent $i$'s best action $a^0_i$ in \cref{alg:MCJESP_initial_best_action}. %
%
%
\Cref{alg:MCJESP_start_node} then creates a start node $n_0$ with $b^0_{BR}$, $a^0_i$, and a weight $w = 1$.
This start node is added to the FSC under construction $(N)$ and an open list $(L)$.

Now, while $L$ is not empty, its node $n$ with largest weight $w$ is popped out (\cref{alg:MCJESP_PopNode}), so as to first develop the nodes that may have the highest impact on the value at the root.
Expanding it requires mapping each observation $o_i$ feasible when performing $n.a_i$ from $n.b$ to a particle set.
This is achieved through sampling by \ProcessAction
until each feasible $o_i$ (according to the samples) is attached to at least $N_{min\mhyphen part}$ particles (\cref{alg:MCJESP_ProcessActionStop}),
which returns a set $\Omega'_i$ of feasible observations, and a mapping $B'_i$ from these observations to particle sets.
%
Then, for each individual observation $o_i$, the algorithm needs to create a transition to an appropriate node, which may already exist
%
%
or needs to be created, as explained in the following.
If $o_i$ is assumed impossible when performing $n.a_i$ in $n.b$ ($o_i \not\in n.\Omega_i$), then a self-loop is added (\cref{alg:MCJESP_selfloop}).\footnote{Note that $o_i$ could become feasible due to future changes in other agents' FSCs.}
Otherwise, %
\cref{alg:MCJESP_BeliefUpdate} gets the belief $b'_{BR}$ attached to $o_i$, and %
\cref{alg:MCJESP_NewWeight} computes an associated weight $w'$. %
%
If %
(i) a belief $\epsilon$-close to $b'_{BR}$ (in 1-norm) exists in $N$, or %
(ii) the FSC has reached its size limit $N_{max\mhyphen\fsc}$ (\cref{alg:MCJESP_CheckNewNode}), %
then we take as next node $n'$ the one in the FSC minimizing $\|b'_{BR} - n'.b\|_1$ and update its weight (\linesref{alg:MCJESP_FindClosestNode}{alg:MCJESP_UpdateClosestNodeWeight}).
Otherwise a next node $n'$ is created using an action selected by POMCP (\linesref{alg:MCJESP_NewBestAction}{alg:MCJESP_BuildNewNode}), and %
added to both $N$ and $L$ (\linesref{alg:MCJESP_BuildNewNode}{alg:MCJESP_addNewNodeToG}).
In \cref{alg:MCJESP_transitionNewNode}, whatever the origin of $n'$, an edge $n \to n'$ is created in the FSC with a label $o_i$.

Note that, for a fixed $N_{max\mhyphen\fsc}$ value, a small $\epsilon$ may prevent from representing long trajectories, while a large $\epsilon$ may induce excessive node merging.

\begin{algorithm}
  \caption{Compute agent $i$'s FSC}
  \label{alg:MCJESP_BuildFSC}
  \DontPrintSemicolon
  \SetInd{.3em}{.6em}

  {[Input:]} %
  $b^0_{BR}$: $G_{BR}$'s initial (extended) belief state $\mid$ %
  \linebreak %
  $G_{BR}$: best response generative model for agent $i$ \; %
  {[Parameters:]}  %
  $N_{max\mhyphen\fsc}$: max FSC size for agent $i$  $ \mid$ %
  \linebreak %
  $N_{min\mhyphen part}$: min number of particles in each belief $\mid$ %
  \linebreak %
  $\epsilon$: min. distance between beliefs  \;
  \Fct{\bf ComputeFSC($b^0_{BR}, G_{BR}$)}{ 
  $a^0_i \gets POMCP(b^0_{BR}, G_{BR})$ \label{alg:MCJESP_initial_best_action} \;
  $n_0 \gets node(b^0_{BR}, a^0_i, w=1)$   \label{alg:MCJESP_start_node} \;
  $N \gets  \{n_0\}$  \tcp*[f]{init FSC \& open list} \;
  $L[w] \gets n_0 $  \tcp*[f]{(sorted by $\searrow$ weight)}\;
  \While{$|L| > 0$ }{
    $n \gets L.popfront()$  \label{alg:MCJESP_PopNode}  \;
    $\Omega'_i, B'_i  \gets \ProcessAction(n.b, G_{BR}, n.a_i)$ \;
    \For(\tcp*[h]{For each obs. of $i$:}){$ o_i \in \Omega_i$}{
      \uIf(\tcp*[h]{$o_i$ unexpected:}){$ o_i \not\in \Omega'_i $}{
        $\eta(n, o_i) \gets n$ \label{alg:MCJESP_selfloop}
        \tcp*[h]{add self-loop}\;
      }
      \Else(\tcp*[h]{Else: create next node}){
        $b'_{BR} \gets B'_{i}[o_i]$ \label{alg:MCJESP_BeliefUpdate} \;
        $w' \gets \frac{|B'_{i}[o_i]|}{|B'_{i}|} \cdot n.w$  \label{alg:MCJESP_NewWeight} \;
        \uIf(\; \tcp*[h]{Similar node exists in FSC or FSC full?}\;
        \tcp*[h]{Yes: Merge with closest node in FSC}){
          $( b'_{BR} \in N(\epsilon) ) \vee ( |N| = N_{max\mhyphen \fsc} ) $ \label{alg:MCJESP_CheckNewNode} }{
          $n' \gets N.findClosest(b'_{BR})$ \label{alg:MCJESP_FindClosestNode}  \;
          $n'.w \gets   n'.w  + w' $ \label{alg:MCJESP_UpdateClosestNodeWeight} \;
        }
      \Else(\tcp*[h]{No: Add new node}) {
          $a'_i \gets POMCP(b'_{BR}, G_{BR} )$ \label{alg:MCJESP_NewBestAction} \;
          $n' \gets node(b'_{BR}, a'_i, w')$  \label{alg:MCJESP_BuildNewNode}  \;
          $N \gets N \cup \{n'\}$\;
          $L[w'] \gets n' $ \label{alg:MCJESP_addNewNodeToG}\;
        }

        $\eta(n, o_i) \gets n'$ \tcp*[h]{Add transition to FSC.} \label{alg:MCJESP_transitionNewNode} \;
      }
    }
  }
}

  \Fct{\ProcessAction{$b^t_{BR}, G_{BR}, a^t_i$}}{
       $\Omega^{t+1}_i \gets \emptyset $ \;
       $B^{t+1}_i \gets \emptyset$ \;
       \Repeat{Timeout() $\vee$ $(MinBeliefParticles(B^{t+1} ) >  N_{min\mhyphen part})$ }{ \label{alg:MCJESP_ProcessActionStop}
       	$e^t \sim b^t_{BR}$ \;
	$\langle e^{t+1}, o^{t+1}_i, r^{t+1}  \rangle \sim G_{BR}(e^t, a^t_i) $ \;
	\If{$ o^{t+1}_i \notin \Omega^{t+1}_i $}{
		 $\Omega^{t+1}_i \gets \Omega^{t+1}_i \cup \{  o^{t+1}_i \} $ \;
	}
	$B^{t+1}[o^{t+1}_i] \gets B^{t+1}[o^{t+1}_i]  \cup \{ e^{t+1} \} $ \;
       }
       \Return{$\Omega^{t+1}_i$, $B^{t+1}_i$}
     }


\end{algorithm}

\subsection{Heuristic Initialization}
\label{sec:MCJESP_init}

Although MC-JESP monotonically improves the value of the joint policy at each iteration, random initializations often lead to poor local optima.
To benefit from a heuristic initialization that allows finding good solutions quickly and reliably,
we adapt \infJESP[]'s heuristics as we adapted the computation of an agent's FSC in the previous section: using particle sets as beliefs, calling a simulation-based solver, and bounding the number of nodes.
In addition, to derive the next belief, we marginalize over possible joint observations $o_{\neq i}$, rather than reasoning on them separately as \cite{InfJESP} did (\eg, considering only the most probable one).



This heuristic initialization assumes that
\begin{enumerate*}
\item agent $i$'s decisions are made as if all agents where sharing their observations, thus acting as a single agent; %
  this means making decisions (picking joint actions $\va$) by solving a Multi-agent POMDP (MPOMDP)  \citep{Pynadath-jair02} relaxation of the original Dec-POMDP, %
  which can be done here with an (online) simulation-based POMDP solver; and 
\item agent $i$'s belief $b$ over the hidden state is updated assuming %
  \begin{enumerate*}
  \item that the other agents ($\neq i$) also act according to the computed MPOMDP policy at $b$, but %
  \item using only $i$'s observation, $o_i$, while marginalizing over others' observations ($o_{\neq i}$, which are actually not known to $i$ at execution time) to ignore them.
  \end{enumerate*}
\end{enumerate*}
%
%
This \textit{one-sided-observation belief update} is computed as follows:
\begin{align*}
  & b^{\va,o_i}(s') \eqdef Pr(s' | o_i, \va, b)
  = \frac{Pr( s', o_i, \va, b)}{Pr( o_i, \va, b)} \\
  & \qquad = \frac{\sum_{\vo_{\neq i}} O( \langle o_i, \vo_{\neq i} \rangle | \va, s') \sum_{s} T( s, \va, s') b(s) }{ \sum_{s', \vo_{\neq i}} O( \langle o_i, \vo_{\neq i} \rangle | \va, s') \sum_s  T(s, \va, s') b(s)} .
\end{align*}

Following this idea, we derive
the FSC heuristic initialization process for agent $i$ detailed in
\Cshref{alg:MCJESP_heuristic} which,
as shown in red, differs from \Cshref{alg:MCJESP_BuildFSC} in two aspects:
\begin{itemize}
\item the Dec-POMDP simulator $G$ is used as an MPOMDP simulator for POMCP to get the best joint action with a given belief (\linesrefAnd{alg:MCJESP_heuristic_initial_best_action}{alg:MCJESP_heuristic_NewBestAction}); and
  \item in \cref{alg:MCJESP_heuristic_process_action}'s \ProcessAction function, the next estimated beliefs are obtained by repeatedly sampling transitions using the computed joint action $\langle n.a_i, n.\va_{\neq i} \rangle$ (and Dec-POMDP simulator $G$), and collecting particle sets for each encountered individual observation $o_i$, ignoring $\vo_{\neq i}$, which is equivalent to a marginalization.
%
\end{itemize}

\begin{algorithm}
  \caption{Build a heuristic FSC for agent $i$}
  \label{alg:MCJESP_heuristic}
  \DontPrintSemicolon
  \SetInd{.3em}{.6em}

  {[Input:]}
  $b^0$: initial state distribution $\mid$
  $i$: agent index  \; 
  {[Parameters:]}
  $G$: Dec-POMDP simulator $\mid$ %
  \linebreak %
  $N_{max\mhyphen \fsc}$: max. FSC size $\mid$ %
  $\epsilon$: min. distance between beliefs \; 

$ \textcolor{red}{\langle a^{0}_{i}, \va^{0}_{\neq i} \rangle   \gets POMCP(b^0, G)} $ \label{alg:MCJESP_heuristic_initial_best_action} \;
  $n_0 \gets node(b^0, a^0_i, \va^{0}_{\neq i}, w=1)$   \label{alg:MCJESP_heuristic_start_node}  \;
  $N \gets  \{n_0\}$  \tcp*[f]{init FSC \& open list} \;
  $L[w] \gets n_0 $ \;
  \While(\tcp*[f]{loop over open nodes}){$|L| > 0$ }{
        $L.sort()$ \label{alg:MCJESP_heuristic_sortG} \;
        $n \gets L.popfront()$  \label{alg:MCJESP_heuristic_PopNode} \;
        $ \textcolor{red}{ \Omega'_i, B'_i  \gets \ProcessAction(n.b, G, \langle n.a_{i}, n.\va_{\neq i} \rangle )}$ \label{alg:MCJESP_heuristic_process_action} \;
        \For(\tcp*[h]{For each obs. of $i$:}){$ o_i \in \Omega$}{
          \uIf(\tcp*[h]{$o_i$ unexpected: add self-loop}){$ o_i \not\in \Omega'_i $}
          {
            $\eta(n, o_i) \gets n$ \label{alg:MCJESP_heuristic_selfloop}
          }
          \Else(\tcp*[h]{Else: create next node:}){
            $b' \gets B'_{i}[o_i]$ \;
            $w' \gets n.w * \frac{|B'_{i}[o_i]|}{|B'_{i}|}$ \;
            %
               \uIf(\; \tcp*[h]{Similar node exists in FSC or FSC full?}\;
              \tcp*[h]{Yes: Merge with closest node in FSC}){
          $( b' \in N(\epsilon) ) \vee ( |N| = N_{max\mhyphen \fsc} ) $ \label{alg:MCJESP_heuristic_CheckNewNode} }{
          $n' \gets N.findClosest(b')$ \label{alg:MCJESP_heuristic_FindClosestNode}  \;
          $n'.w \gets   n'.w  + w' $ \label{alg:MCJESP_heuristic_UpdateClosestNodeWeight} \;
        }
      \Else(\tcp*[h]{No: Add new node}) {
           $ \textcolor{red}{\langle a'_{i}, \va'_{\neq i} \rangle   \gets POMCP(b', G)} $ \label{alg:MCJESP_heuristic_NewBestAction} \;
              $n' \gets node(b', a'_i, \va'_{\neq i}, w')$  \label{alg:MCJESP_heuristic_BuildNewNode}  \;
              $N \gets N \cup \{n'\}$\;
              $L[w] \gets n' $ \label{alg:MCJESP_heuristic_addNewNodeToG}\;
        }

            $\eta(n, o_i) \gets n'$ \label{alg:MCJESP_heuristic_transitionNewNode} \;
          }
        }
}


\end{algorithm}

\subsection{Observations}


With an increasing time budget, POMCP asymptotically converges to optimal decisions.
By %
(i) increasing POMCP's time budget to infinity, %
(ii) increasing $N_{min\mhyphen part.}$ and $N_{max\mhyphen \fsc}$ to infinity, 
and %
(iii) setting $\epsilon=0$, 
each iteration of the local search would thus return the best response (possibly infinite) FSC.
As a consequence, assuming also an exact evaluation of FSCs, the local search would be guaranteed to find a Nash equilibrium.

In practice, we only obtain approximate Nash equilibria.
Also, due to randomization in POMCP and in FSC evaluations through simulations, restarts of the full process lead to different search trajectories, but always stop in finitely many iterations with probability 1.
The next section looks at the results obtained in practice through experiments.

\section{Experiments}
\label{sec:exp}

We evaluate our contributions on five state-of-the-art Dec-POMDP benchmarks (\cf \url{http://masplan.org/problem_domains}):
%
Decentralized Tiger 
\citep{JESP},
Recycling Robots 
\citep{Amato2007UAI},
Meeting in a $3 \times 3$ grid 
\citep{amato2009incremental},
Cooperative Box Pushing 
\citep{Seuken2007ImprovedMD},
Mars Rover \citep{amato2009achieving}.
%
We compare MC-JESP with state-of-the-art Dec-POMDP solvers relying on either:
%
{\em explicit models:} (which benefit from more information)  FB-HSVI \citep{DibAmaBufCha-jair16}, Peri \citep{PajPel-nips11}, PeriEM \citep{PajPel-nips11}, PBVI-BB \citep{MacIsb-nips13}, MealyNLP \citep{AmaBonZil-aaai10} and \infJESP; or
{\em generative models:} MCEM \citep{Wu2013}, Dec-SBPR \citep{Liu2015} and oSARSA \citep{pmlr-v80-dibangoye18a}.
%

For MC-JESP,
\begin{enumerate*}
\item POMCP is used as our simulation-based POMDP planner with a timeout of $1$\,s;
\item we consider three different maximum FSC sizes: 10, 30, and 50, respectively;
  %
\item the threshold distance between beliefs is set to $\epsilon=0.1$; and
\item FSC evaluations (\cref{codes:PolicyEval_infJESP} of \Cshref{alg:JESP_main}) use $10^6$ simulations that stop when $\gamma^t < 10^{-4}$.
\end{enumerate*}
For MC-JESP and \infJESP's empirical results, having access to the exact model in each benchmark problem, we use \citeauthor{Hansen-nips97}'s [\citeyear{Hansen-nips97}] policy evaluation for FSCs applied to a best-response POMDP.
%
The experiments with MC-JESP were conducted on a laptop with a 2.3 GHz i9 CPU.
The source code 
is in the supplementary material.

\subsection{Comparison with state-of-the-art algorithms}
\label{sec:results}

\begin{table}
  \centering
  \caption{Comparison of different algorithms in terms of final FSC size, number of iterations, time, and value, on 5 infinite-horizon benchmarks with $\gamma = 0.9$ for all domains.
The solvers are listed in a decreasing order of value.
  }
  \label{Table:MCJESP_BenchmarksResults}
  \resizebox{\linewidth}{!}{%
    
\def\InfJesp{IJ}
\def\Rand{R}

\newcommand{\stdv}[1]{{\scriptstyle \pm #1}}

\begin{tabular}{
  S
  c
  S
  S[round-mode = places, round-precision = 0, table-format=2.0]
  S[round-mode = places, round-precision = 2, table-format=2.2]
  }
  \toprule
  {Alg.} & {FSC size} & {Iterations} & {Time (s)} & {Value}\\
  \midrule
  \multicolumn{5}{c}{ {DecTiger} {($|\cI|=2, |\cS|=2, |\cA^{i}|=3, |\cZ^{i}|=2 $)}} \\ 
  \midrule
  {FB-HSVI*} & & & 153.7 & 13.45 \\
    {Peri*} & & & 220 & 13.45  \\
    {INF-JESP*} & {$ 6 \times 6 $} & 27 & 213 & 13.44 \\
    \textbf{MC-JESP(M-{$20$})} & {$ 30 \times 30 $} & 5 & 5620  & 13.44 \\
  {PeriEM*} & & & 6450 & 9.42 \\
   {oSARSA} & & &  & -0.2  \\
  \textbf{MC-JESP(M-{$1_{20}$})} & {$ 24 \times 25 $} & 4 & 281 & -2.33 \\
  {Dec-SBPR} & & & 96 & -18.63  \\
  {MCEM} & & & 20 & -32.31  \\

  \midrule
  \multicolumn{5}{c}{{Recycling} {($|\cI|=2, |\cS|=4, |\cA^{i}|=3, |\cZ^{i}|=2 $)}} \\
  \midrule
  {FB-HSVI*} & & & 2.6 & 31.929 \\
  \textbf{MC-JESP(M-{$20$})} & {$ 17 \times 17 $} & 3 & 5260 & 31.92 \\
  {Peri*} & & & 77 & 31.84  \\
  {PeriEM*} & & & 272 & 31.80 \\
  {INF-JESP*} & {$ 2 \times 2 $} & 3 & 3.1 & 31.62 \\
    {Dec-SBPR} & & & 147 & 31.26  \\
  \textbf{MC-JESP(M-{$1_{20}$})}& {$ 19 \times 20 $} & 4 & 263 & 30.74 \\

  \midrule
  \multicolumn{5}{c}{{Grid3*3} {($|\cI|=2, |\cS|=81, |\cA^{i}|=5, |\cZ^{i}|=9 $)}} \\
  \midrule
  {INF-JESP*} & {$ 8 \times 10 $} & 3 & 2 & 5.81 \\
  \textbf{MC-JESP(M-{$20$})} & {$ 50 \times 50 $} & 4 &  11900 & 5.81 \\
  {FB-HSVI*} & & & 67 & 5.802 \\
  \textbf{MC-JESP(M-{$1_{20}$})}& {$ 50 \times 50 $} & 6 & 595 & 5.80 \\
  {Peri*} & &  & 9714 & 4.64 \\
  \midrule
  \multicolumn{5}{c}{{Box-pushing} {($|\cI|=2, |\cS|=100, |\cA^{i}|=4, |\cZ^{i}|=5 $)}} \\
  \midrule
  {FB-HSVI*} & &  & 1715.1 & 224.43 \\
    \textbf{MC-JESP(M-{$20$})} & {$ 50 \times 50 $} & 6 & 9740  & 223.84 \\
      \textbf{MC-JESP(M-{$1_{20}$})}& {$ 50 \times 50 $} & 5 & 487 & 220.94 \\
   {INF-JESP*} & {$ 250 \times 408 $} & 6 & 963 & 220.25 \\
  {Peri*} & & & 5675 & 148.65 \\
     {oSARSA} & & &  & 144.57  \\
  {PeriEM*} & & & 7164 & 106.65 \\
      {Dec-SBPR} & & & 290 & 77.65  \\

  \midrule
  \multicolumn{5}{c}{{Mars Rover} {($|\cI|=2, |\cS|=256, |\cA^{i}|=6, |\cZ^{i}|=8 $)}} \\
  \midrule
  {FB-HSVI*} & & & 74.31 & 26.94 \\
     {INF-JESP*} & {$ 125 \times 183 $} & 6 & 122 & 26.91\\
    \textbf{MC-JESP(M-{$20$})} & {$ 50 \times 50 $} & 5 & 6980 & 26.45 \\
  \textbf{MC-JESP(M-{$1_{20}$})}& {$ 50 \times 50 $} & 3 & 349 & 25.89 \\
  {Peri*} & & & 6088 & 24.13 \\
    {Dec-SBPR} & & & 1286 & 20.62  \\
  {PeriEM*} & & & 7132 & 18.13 \\ 
  \bottomrule
\end{tabular}

  }
\end{table}

\Cref{Table:MCJESP_BenchmarksResults} presents the results for the 5 problems, the solvers being ordered from best to worse value. 
Among $x$ restarts of MC-JESP, the best value is reported in MC-JESP(M-$x$), and the average value in MC-JESP(M-$1_x$) (to look at the benefit of restarting).
For \infJESP, we report the best values among its 3 possible initializations \citep{InfJESP}.
For MC-JESP, we report the best values over the 3 possible max. FSC sizes.
%
%
The columns provide:
\begin{itemize*}
\item (\textit{Alg.}) the different algorithms at hand, with a $*$ exponent for those who rely on an explicit model;
\item (\textit{FSC size}) the final FSC size  (for \infJESP[s] and MC-JESP);
%
\item (\textit{Iterations}) the number of iterations required to converge (for \infJESP[s] and MC-JESP);
\item (\textit{Time}) the running time;
\item (\textit{Value}) the final value (lower bounds for \infJESP[s] and MC-JESP, the true value being at most 0.01 higher).
\end{itemize*}
In terms of final value achieved, MC-JESP(M-$1_{20}$) finds good solutions in all cases except DecTiger (which is a small but difficult coordination problem), and MC-JESP(M-$20$) obtains  results very close to FB-HSVI's near-optimal solutions, which rely on an explicit Dec-POMDP model, for all benchmark problems.
MC-JESP even sometimes achieves better results than some other explicit model-based algorithms.
Also, compared with other simulation-based methods, it dominates in large problems (Box-pushing and Mars Rovers), while other simulation-based methods fail.

However,  compared with the explicit model-based algorithms, MC-JESP requires more solving time.
For example, in large problems such as Mars Rovers, MC-JESP takes $349$\,s on average to solve the task, while \infJESP takes $122$\,s.
But this is not surprising since MC-JESP only uses a black-box simulator.
%
A key question is how to determine whether restarting can be beneficial.

\subsection{A Closer Look at MC-JESP's behavior}

\begin{figure*}
  \def\scale{0.95}
       \centering

       \includegraphics[width=\scale\columnwidth]{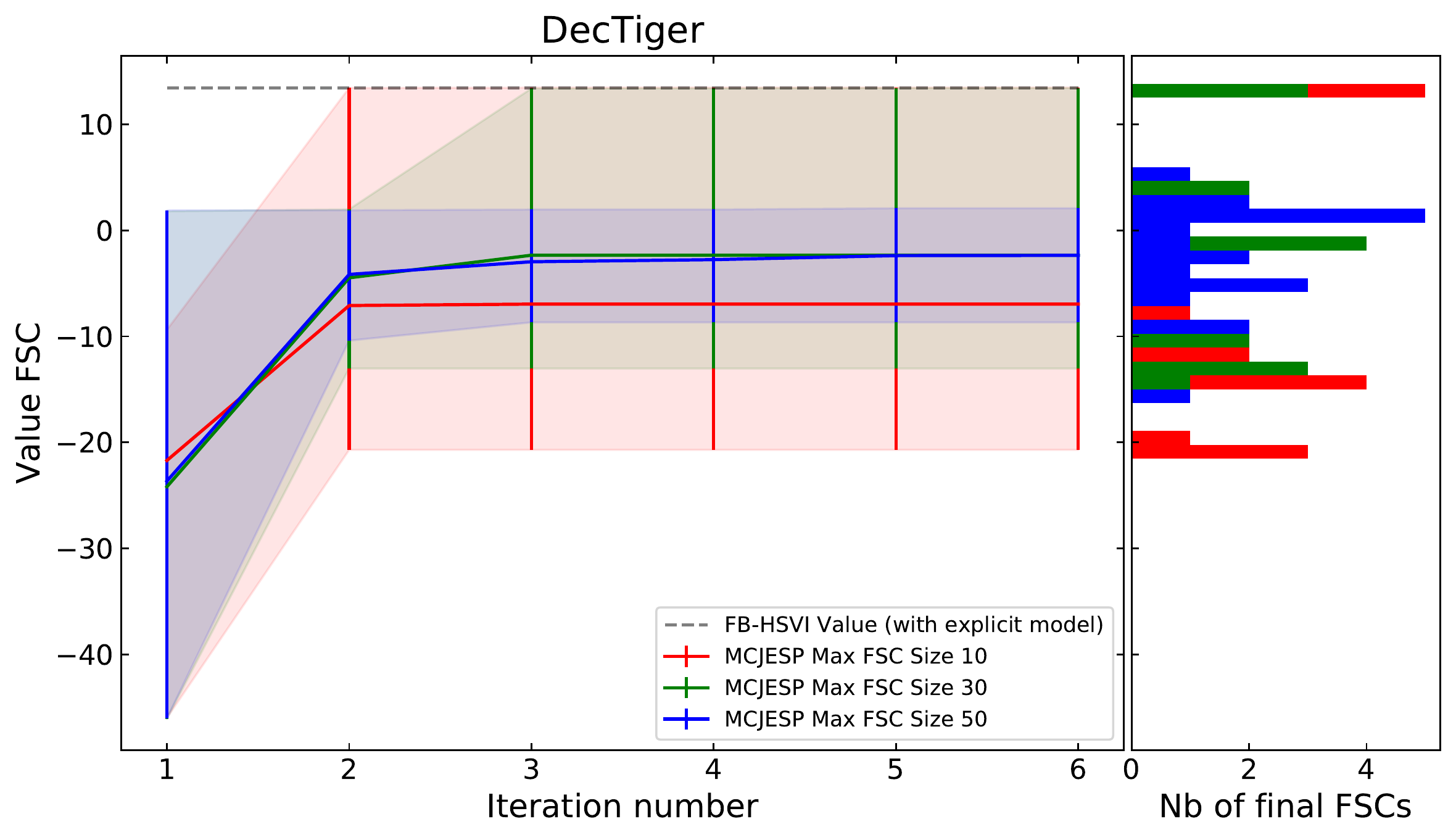}
       \hfill
       \includegraphics[width=\scale\columnwidth]{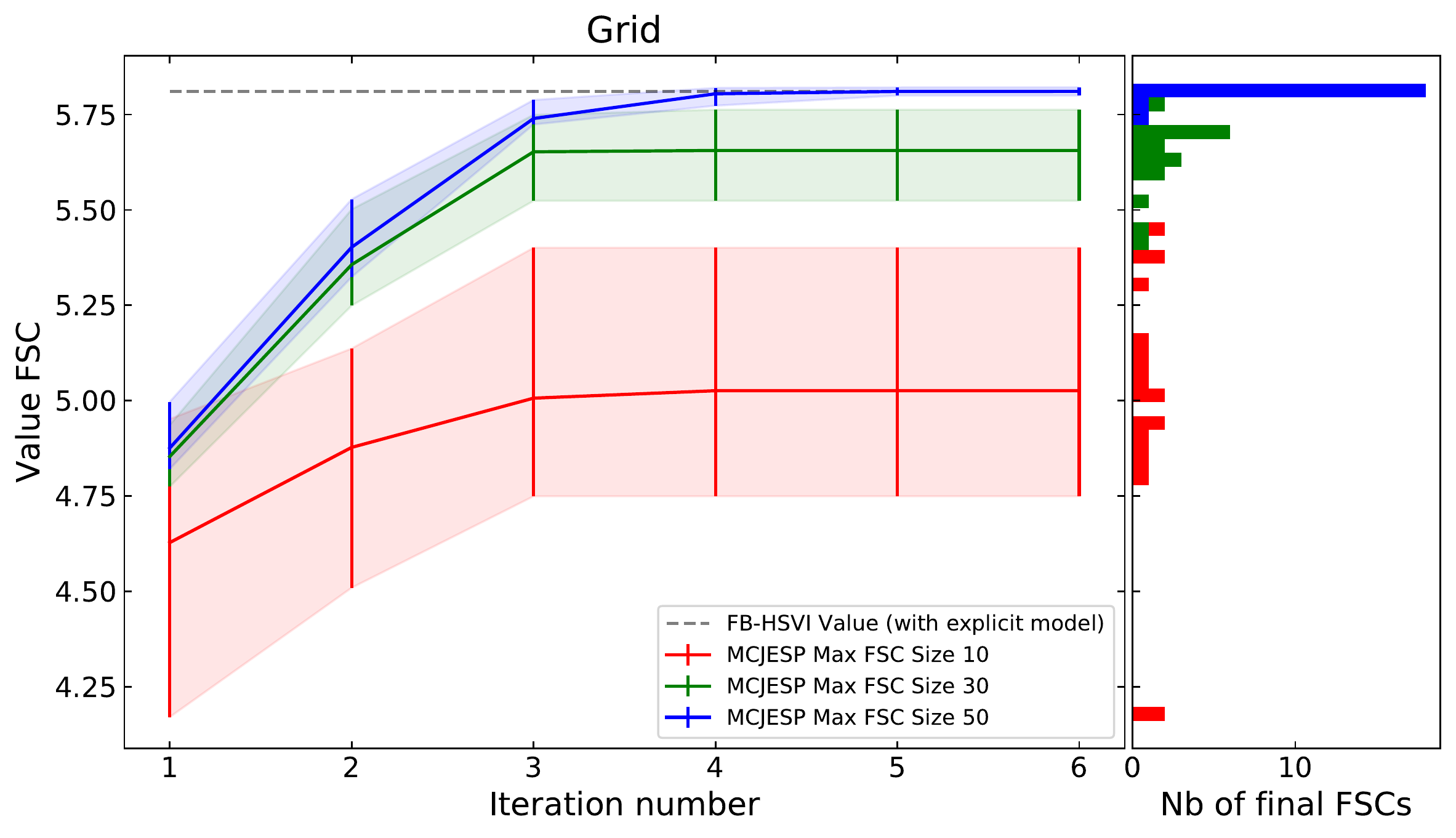}

       \includegraphics[width=\scale\columnwidth]{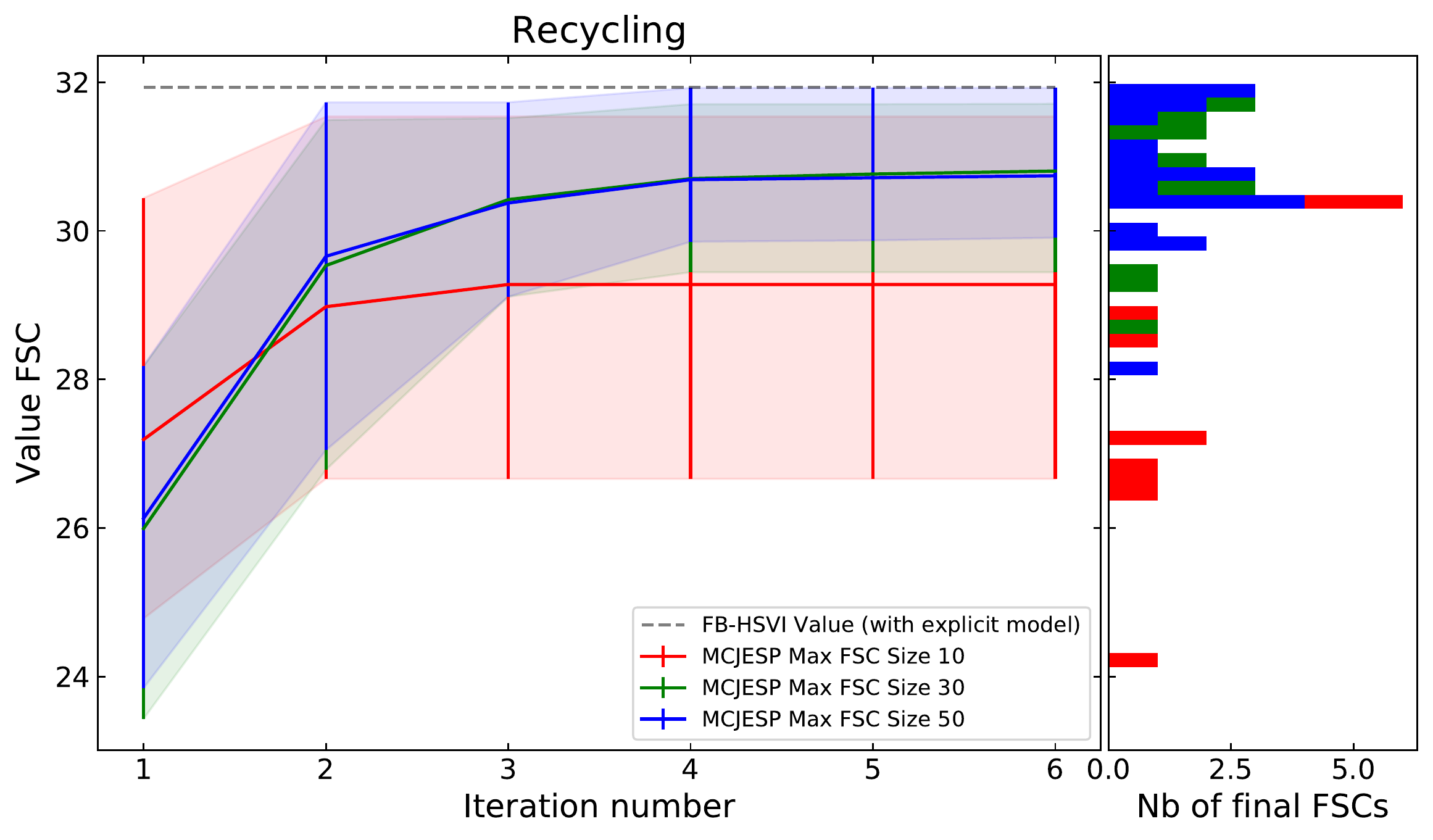}
       \hfill
       \includegraphics[width=\scale\columnwidth]{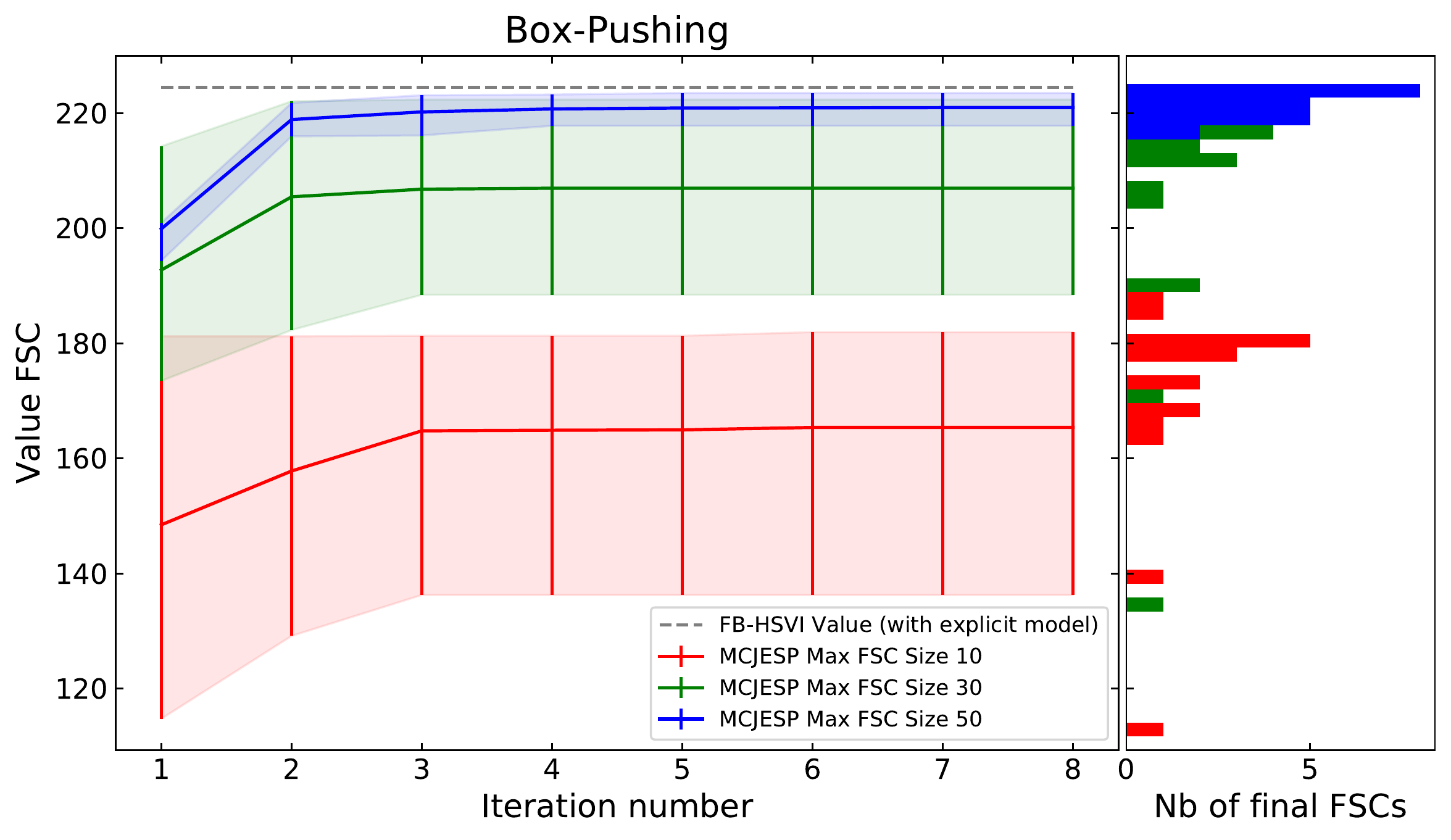}

       \includegraphics[width=\scale\columnwidth]{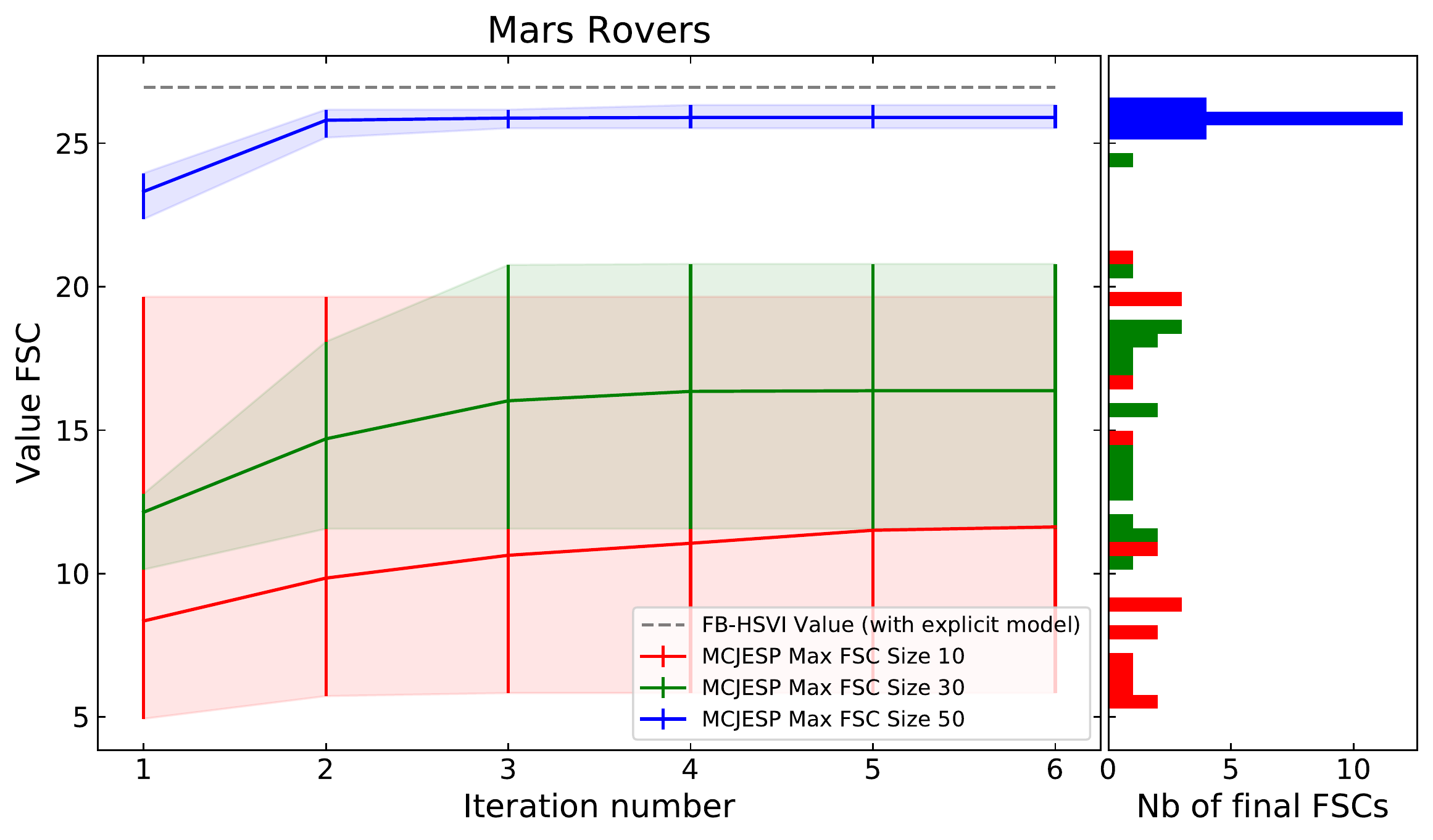}
       \hfill
       \begin{minipage}[b]{\columnwidth}
       \caption{%
         Values of the joint policy for the Dec-Tiger, Grid, Recycling, Box-Pushing, and Mars Rover problems (from top to bottom).
         The left part of each figure presents the evolution (during a
         run) of the value of the joint policy at each iteration of MC-JESP($1_{20}$) (avg + 10th and 90th percentiles) with different bounded FSC sizes (10, 30, and 50, respectively).
       The dashed line represents FB-HSVI's final value.
       The right part presents the value distribution after convergence of MC-JESP($1_{20}$).
     }
     \label{Figure:MCJESP_ValueIterationAndFinal}
     \end{minipage}
\end{figure*}

\begin{figure}
       \centering

       \includegraphics[width=.85\linewidth]{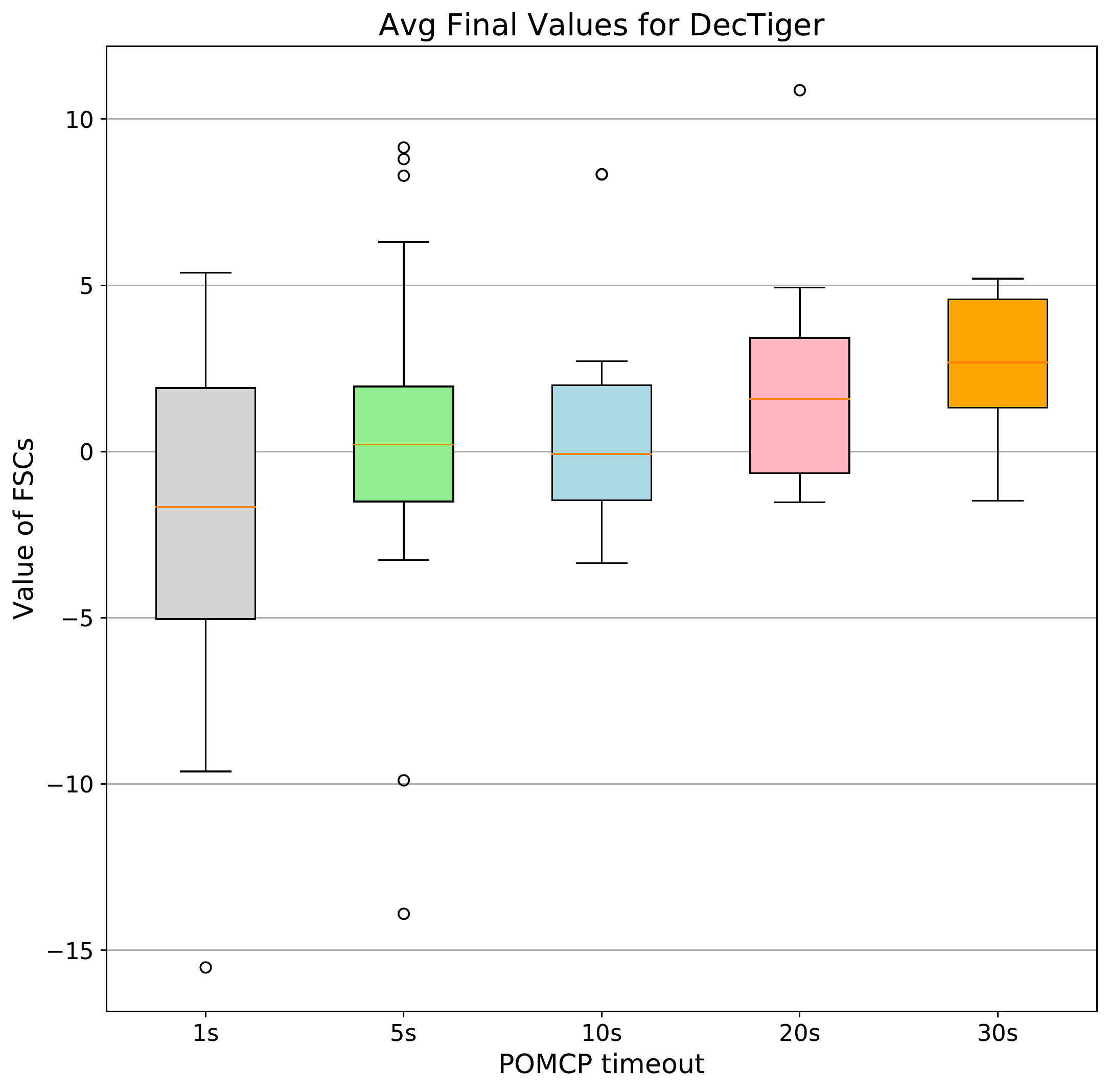}
       \caption{%
         Values of the joint policy for the Dec-Tiger problem for different POMCP timeout values.
     }
     \label{Figure:MCJESP_ImpactTimeout}
\end{figure}


We study MC-JESP's performance with three different maximum FSC sizes in \Cshref{Figure:MCJESP_ValueIterationAndFinal} (red for 10, green for 30, and blue for 50).
%
Right parts present the distribution over final values of MC-JESP with 20 restarts.
In the five problems at hand, MC-JESP with max FSC size 50 (blue) has distributions more concentrated on good values than others, and most values are close to FB-HSVI's ones (thus, near-optimal values).
These distributions show that few restarts are needed to reach good solutions with high probability if we give large enough FSC sizes. 
The left parts of \Cshref{Figure:MCJESP_ValueIterationAndFinal} present the evolution of the values during each iteration of MC-JESP with the three maximum FSC sizes.
The average is computed over all runs, even if they have already converged.
This figure first shows that MC-JESP monotonically increases during each run, and most runs converge to good local optima in a few iterations.
Second, we observe that, for large problems (Box-Pushing and Mars Rovers), there are already significant drops from MC-JESP in the first iteration with an FSC size limit decreasing from 50 to 10.
This indicates that, for large problems, we must give large enough FSC size limits, while this is not necessary for small problems.

Last but not least, in Dec-Tiger, although some restarts of MC-JESP end with optimal values, we observe that the average value is still relatively low compared with FB-HSVI.
Therefore, we conducted another experiment to investigate the impact of different POMCP timeouts (note that there is a fixed timeout of $1$\,s for the experiments illustrated in \Cref{Figure:MCJESP_ValueIterationAndFinal}).
To that end, we limit the FSC size in each iteration to at most 50 nodes, and we test MC-JESP with five POMCP timeouts (1\,s,  5\,s, 10\,s, 20\,s, and 30\,s).
The distribution of final values is shown in \Cshref{Figure:MCJESP_ImpactTimeout}.
We observe that the average value increases and the variability shrinks when we give more time to POMCP.
However, it also indicates that, when we increase the time budget,  we have a lower chance of getting "lucky" good values.
%


\section{Conclusion}
\label{sec:conclusion}

In this work, based on \infJESP, we propose a novel infinite-horizon Dec-POMDP solver called MC-JESP,  which only requires a black-box Dec-POMDP simulator, and returns FSCs, \ie, representations that can make for interpretable policies.
We describe how to obtain a best-response generative model (the simulator of the POMDP faced by some agent $i$ assuming known FSCs for other agents), and the process to extract an FSC for each agent.
Moreover, a heuristic initialization method for MC-JESP is also provided.

Through experiments, we prove that MC-JESP preserves \infJESP[]'s competitive results (though at the cost of an increased computation time), performing better than many explicit model-based algorithms, and outperforming other simulation-based algorithms in most cases.
Because it seeks Nash equilibria, this approach could better scale up to large problems than approaches directly seeking global optima.



Several improvements of MC-JESP could be envisioned, such as:
\begin{enumerate*}
\item robustly comparing FSCs $\fsc_i$ and $\fsc'_i$, while minimizing computation time through hypothesis testing; %
\item if using large FSCs, using space partitioning (\eg, $k$-d trees \citep{10.1145/361002.361007} or cover trees \citep{beygelzimer2006cover}) to speed up the search for nearest nodes; and %
\item
%
re-using POMCP trees from one node to the next, or to initialize \ProcessAction, although doing so may significantly increase memory usage. %
\end{enumerate*}

Also, preliminary experiments show that MC-JESP works on a continuous-state meet-in-a-grid problem, the main issue being to replace the distance between sets of discrete particles (i.e., just comparing two vectors representing discrete distributions)
by a distance over continuous particles (which requires taking the distance between states into account).
%
%
For future works, we plan to extend MC-JESP to problems with continuous actions and observations.
This would require not only relying on algorithms such as \citeauthor{sunberg2018online}'s POMCPOW [\citeyear{sunberg2018online}], but also, more importantly, deriving FSCs that can handle continuous observations.
%

\bibliography{MCJESP}
\end{document}